%% file: main.tex
\documentclass[10pt,twocolumn,letterpaper]{article}

\usepackage[pagenumbers]{cvpr} 

\usepackage{graphicx}
\usepackage{amsmath}
\usepackage{amssymb}
\usepackage{booktabs}
\usepackage{xcolor}
\usepackage[pagebackref,breaklinks,colorlinks]{hyperref}

\usepackage[capitalize]{cleveref}
\crefname{section}{Sec.}{Secs.}
\Crefname{section}{Section}{Sections}
\Crefname{table}{Table}{Tables}
\crefname{table}{Tab.}{Tabs.}

\newlength\savewidth\newcommand\shline{\noalign{\global\savewidth\arrayrulewidth
  \global\arrayrulewidth 1pt}\hline\noalign{\global\arrayrulewidth\savewidth}}
\usepackage{pifont}

\begin{document}

\title{Real-time Neural Radiance Talking Portrait Synthesis \\ via Audio-spatial Decomposition}

\author{
Jiaxiang Tang$^1$\thanks{\noindent Work done while at Baidu Inc.} \quad
Kaisiyuan Wang$^3$ \quad
Hang Zhou$^2$ \quad
Xiaokang Chen$^1$ \quad
Dongliang He \quad \\
Tianshu Hu$^2$ \quad
Jingtuo Liu$^2$ \quad
Gang Zeng$^1$ \quad
Jingdong Wang$^2$ \\
$^1$School of Intelligence Science and Technology, Peking University \\
\quad $^2$Baidu Inc. \quad $^3$The University of Sydney \\
{\tt\small \{tjx, pkucxk, zeng\}@pku.edu.cn} \quad {\tt\small kaisiyuan.wang@sydney.edu.au} \quad {\tt\small hedlcc@126.com}\\
{\tt\small \{zhouhang09,hutianshu01,liujingtuo,wangjingdong\}@baidu.com}
}
\maketitle

\begin{abstract}
\input{abstract}
\end{abstract}

\section{Introduction}
\input{intro}

\section{Related Work}
\input{related}

\section{Method}
\input{method}

\section{Experiment}
\input{experiments}

\section{Broader Impact}
\input{impact}

\section{Conclusion}
\input{conclusion}

{\small
\bibliographystyle{ieee_fullname}
\bibliography{refs}
}

\input{appendix.tex}

\end{document}

%% file: abstract.tex
While dynamic Neural Radiance Fields (NeRF) have shown success in high-fidelity 3D modeling of talking portraits, the slow training and inference speed severely obstruct their potential usage. 
In this paper, we propose an efficient NeRF-based framework that enables real-time synthesizing of talking portraits and faster convergence by leveraging the recent success of grid-based NeRF.  
Our key insight is to decompose the inherently high-dimensional talking portrait representation into three low-dimensional feature grids. 
Specifically, a Decomposed Audio-spatial Encoding Module models the dynamic head with a 3D spatial grid and a 2D audio grid.
The torso is handled with another 2D grid in a lightweight Pseudo-3D Deformable Module.
Both modules focus on efficiency under the premise of good rendering quality.
Extensive experiments demonstrate that our method can generate realistic and audio-lips synchronized talking portrait videos, while also being highly efficient compared to previous methods.

%% file: intro.tex
Audio-driven talking portrait synthesis is a long-standing task with various applications, such as digital human creation, virtual video conferencing, and film-making. 
Previously, researchers have explored various approaches to solve this problem.
Some studies focus on driving a single or a few images~\cite{chung2017you,zhou2019talking,pham2017speech,prajwal2020lip,chen2019hierarchical,cudeiro2019capture,chen2020talking,das2020speech,zhou2020makelttalk,zhou2021pose,liang2022expressive,ji2022eamm}, which mostly create non-realistic results.
Other methods model a specific person by leveraging explicit facial structural priors such as landmarks and meshes~\cite{suwajanakorn2017synthesizing,thies2020neural,ji2021audio,song2022everybody}, but the errors accumulated in predicting such intermediate representations would greatly affect the final results.

Neural Radiance Fields (NeRF)~\cite{mildenhall2020nerf} has recently received broad attention for its capability in rendering realistic and 3D consistent novel view images, which is also essential in talking portrait synthesis. 
NeRFace~\cite{gafni2021dynamic} firstly involves 3DMM parameters in the control of a human head NeRF. 
Meanwhile, several studies~\cite{guo2021ad,liu2022semantic,shen2022learning} exploit the audio-driven setting by directly conditioning NeRF on audio signals without intermediate representations.
Though success has been shown in their portrait synthesis results, they all suffer from one major drawback: the \emph{slow speed} in video production.
For example, it takes AD-NeRF~\cite{guo2021ad} 12 seconds to render a $450 \times 450$ frame on an RTX 3090 GPU, which is far from real-time applicable (25 FPS).
The training also takes days for each target person, limiting the wider applications of such methods. 

To tackle this problem, an intuitive idea is to take inspiration from recent advances in improving the efficiency of NeRF~\cite{mueller2022instant,TensoRF,sun2021direct,yu_and_fridovichkeil2021plenoxels,yu2021plenoctrees,liu2020neural,chan2021efficient}. 
The common idea behind these works is to reduce the MLP size and store 3D scene features in an explicit trainable grid structure.
The expensive MLP forwarding is then replaced by cheaper linear interpolations to retrieve desired information at each static 3D location.
However, these static scene-oriented methods are not directly applicable to our dynamic case.

We identify \emph{two major challenges} in building real-time NeRF-based audio-driven portraits:
\textbf{1)} 
How to efficiently represent both spatial and audio information with grid-based NeRF remains unsolved.
Normally, audios are encoded to a $64$ dimension vector~\cite{guo2021ad,liu2022semantic} and fed into the MLP with 3D spatial coordinates.
However, involving additional dimensions of audio in grid-based settings for linear interpolation would lead to exponential computational complexity growth.
\textbf{2)} Efficient modeling of the less complicated but equally essential torso part for realistic portraits is not trivial.
The previous practices either involve another full 3D radiance field~\cite{guo2021ad} or learn an entangled 3D deformation field~\cite{liu2022semantic}, which are excessive and expensive.

In this work, we present a novel framework called \textbf{Real-time Audio-spatial Decomposed NeRF (RAD-NeRF)}, which allows efficient training and real-time inference for audio-driven talking portraits.
We take the advantage of a recent grid-based NeRF representation~\cite{mueller2022instant} and adapt its efficient static scene modeling capability to dynamic audio-driven portrait modeling.
Our key insight is to \emph{explicitly decompose the inherently high-dimensional audio-guided portrait representation into three low-dimensional trainable feature grids}.
Specifically, for dynamic head modeling, we propose a Decomposed Audio-spatial Encoding Module which decomposes audio and spatial representations into two grids.
While we keep the static spatial coordinates in 3D, the audio dynamics are encoded as low-dimensional ``coordinates''.
Moreover, instead of querying audio and spatial coordinates within one higher-dimensional feature grid, 
we show that they can be divided into two separate lower-dimensional feature grids, which further reduces the cost of interpolation. 
This decomposed audio-spatial encoding enables an efficient dynamic NeRF for talking head modeling.

As for the torso part, we look into its motion patterns in pursuit of lower computational costs. 
Given the observation that topological changes are less involved in torso movements, we propose a lightweight Pseudo-3D Deformable Module to model the torso with a 2D feature grid.
Combining these two modules with further portrait-specific NeRF acceleration designs, our method can achieve real-time inference speed with a modern GPU.

Our contributions can be summarized as follows: 
\begin{itemize}
\item We propose a Decomposed Audio-spatial Encoding Module to efficiently model the inherently high-dimensional audio-driven facial dynamics with two low-dimensional feature grids.
\item We propose a lightweight Pseudo-3D Deformable Module to further enhance efficiency in synthesizing natural torso motion synchronized with the head motion.
\item Our framework can run 500$\times$ faster than the previous works with better rendering quality, and also supports various explicit controls of the talking portrait such as head pose, eye blink, and background image.
\end{itemize}

%% file: related.tex
\begin{figure*}[t!]
    \centering
    \includegraphics[width=\textwidth]{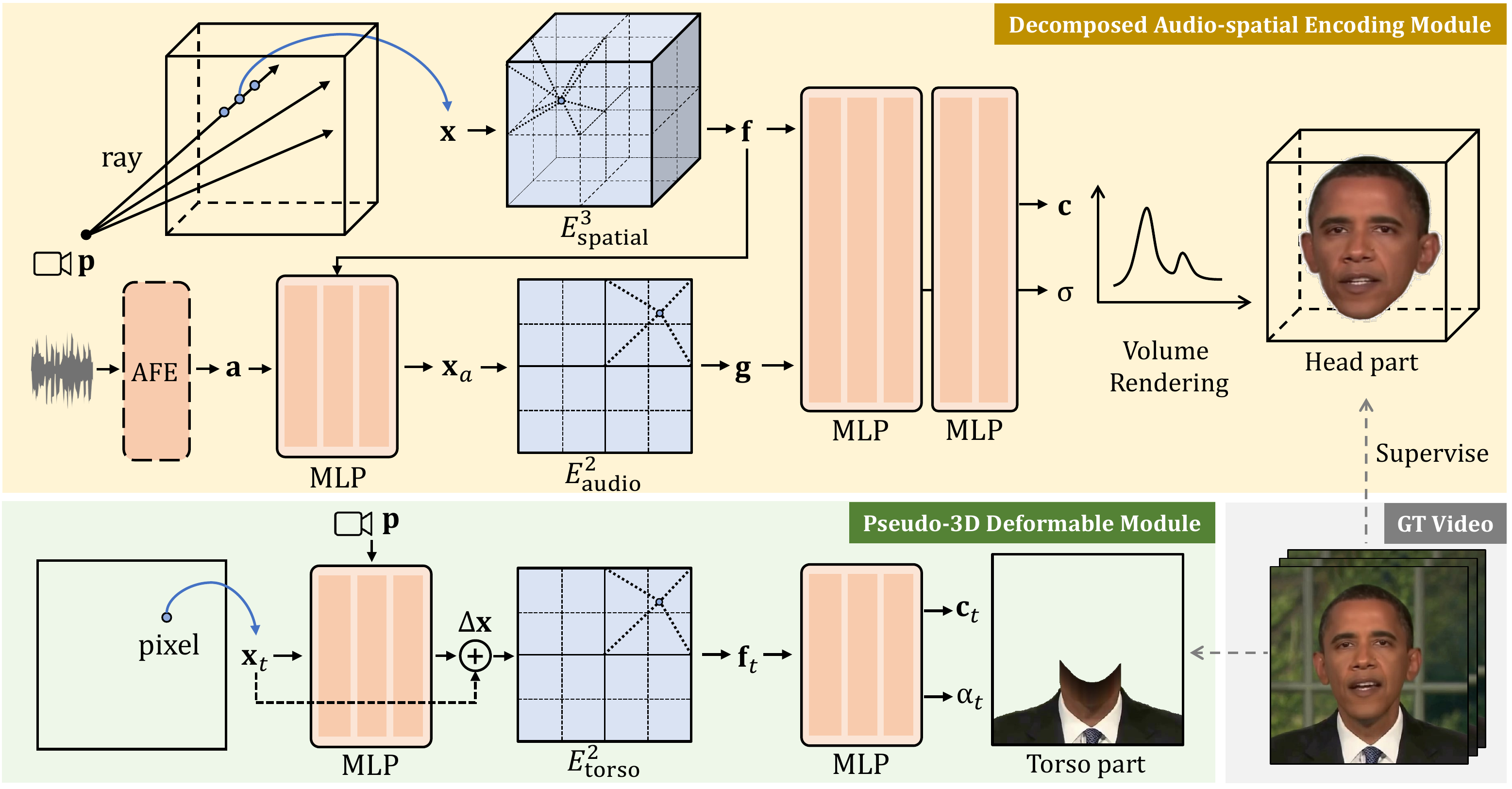}
    \caption{\textbf{Network Architecture}.
    The head is modeled with the Audio-spatial Decomposed Encoding Module.
    Input audio signal is first processed with the Audio Feature Extractor (AFE)~\cite{guo2021ad}, then compressed to a low-dimensional spatial-dependent audio coordinate $\mathbf{x}_a$.
    Two decomposed grid encoders $E^3_\text{spatial}, E^2_\text{audio}$ separately encode the spatial coordinate $\mathbf{x}$ and audio coordinate $\mathbf{x}_a$.
    The spatial features $\mathbf{f}$ and audio features $\mathbf{g}$ are fused in an MLP to produce head color $\mathbf{c}$ and density $\sigma$ for volume rendering.
    The torso is modeled with the Pseudo-3D Deformable Module.
    We only sample one torso coordinate $\mathbf{x}_t$ per pixel, and learn a deformation field to model the torso dynamics dependent on head pose $\mathbf{p}$.
    Another grid encoder $E^2_\text{torso}$ learns the torso features $\mathbf{f}_t$, which are fed to an MLP to get torso color $\mathbf{c}_t$ and alpha $\alpha_t$.
    }
    \vspace{-0.3cm}
    \label{fig:network}
\end{figure*}

\subsection{Audio-driven Talking Portrait Synthesis}
Audio-driven talking portrait synthesis aims to reenact a specific person given arbitrary input speech audio.
Various approaches have been proposed to achieve realistic and well-synchronized talking portrait videos.
Conventional methods~\cite{bregler1997video,brand1999voice} define a set of phoneme-mouth correspondence rules and use stitching-based techniques to modify mouth shapes.
Deep learning enables image-based methods~\cite{chung2017you,cudeiro2019capture,ji2021audio,pham2017speech,prajwal2020lip,song2018talking,taylor2017deep,vougioukas2020realistic,wang2021audio2head,zhang2021flow,zhou2019talking,zhou2021pose} by synthesizing images corresponding to the audio inputs.
One limitation of these methods is that they can only generate images at a fixed resolution, and cannot control the head poses.
Another line of research focuses on model-based methods~\cite{chen2020talking,chen2019hierarchical,das2020speech,meshry2021learned,song2022everybody,suwajanakorn2017synthesizing,thies2020neural,wang2020mead,wu2021imitating,yi2020audio,zhou2020makelttalk,Gao2022nerfblendshape}, where structural representations like facial landmarks and 3D morphable face models are used to aid the talking portrait synthesis. 
However, the estimation of these intermediate representations may introduce extra errors.
Recently, several works~\cite{gafni2021dynamic,guo2021ad,liu2022semantic,shen2022learning} leverage NeRF~\cite{mildenhall2020nerf} to synthesize talking portraits.
NeRF-based methods can achieve photorealistic rendering at arbitrary resolution with less training data, but current works on audio-driven talking portrait synthesis still suffer from slow training and inference speed.

\subsection{Neural Radiance Field}
NeRF~\cite{mildenhall2020nerf} combines implicit neural representation with volume rendering for novel view synthesis, achieving unprecedented photorealistic results.
This powerful representation has received broad attention and is studied in various topics. 

\noindent \textbf{Dynamic Modeling.}
Since the vanilla NeRF is only capable of modeling static scenes, many distinct approaches have been proposed for modeling dynamic scenes~\cite{pumarola2021d,park2021nerfies,park2021hypernerf,gafni2021dynamic,guo2021ad,li2021neural,tretschk2021non,peng2021neural,peng2021animatable}. 
In particular, \textit{Deformation-based} methods~\cite{pumarola2021d,park2021nerfies} aim to map all observations back to a canonical space, by learning a deformation field along with the radiance field. 
\textit{Modulation-based} methods~\cite{park2021nerfies,gafni2021dynamic,guo2021ad,liu2022semantic,park2021hypernerf} directly condition NeRF on a latent code, which can represent the time or audio. 
These methods are better for modeling complex dynamics involving topological changes, and are more suitable for modeling facial dynamics.

\noindent \textbf{Efficiency.}
Enhancing the efficiency of the NeRF representation is of great importance to its practical applications. 
The vanilla NeRF and most following works~\cite{mildenhall2020nerf,barron2021mip,barron2021mipnerf,park2021nerfies,gafni2021dynamic,guo2021ad,liu2022semantic} only rely on one neural network, typically a Multi-layer Perceptron (MLP), to encode the 3D scenes.
This type of \textit{implicit} NeRF suffers from slow training and inference speed, since rendering an image requires tremendous evaluations of the MLP at all sampled coordinates.
In order to accelerate rendering, recent works~\cite{liu2020neural,yu2021plenoctrees,yu_and_fridovichkeil2021plenoxels,sun2021direct,mueller2022instant,TensoRF,tang2022compressible} propose to reduce the size of the MLP or totally remove it, and store the 3D scene features in an explicit 3D feature grid structure.
For example, DVGO~\cite{sun2021direct} directly uses a dense feature grid for acceleration. 
Instant-NGP~\cite{mueller2022instant} adopts a multi-resolution hash table to control the model size.
TensoRF~\cite{TensoRF} factorizes the dense 3D feature grid into compact low-rank tensor components.
However, these \emph{grid-based} NeRF are only applicable to static scenes.
Several works~\cite{fang2022fast,guo2022neural,wang2022fourier,liu2022devrf} apply these acceleration techniques to dynamic NeRF, but are either deformation-based or only support time-dependent dynamics, which are not suitable for audio-driven talking portrait synthesis.
In contrast, our method is specially designed for the audio-driven setting in talking portrait synthesis.

%% file: method.tex
In this section, we present our Real-time Audio-spatial Decomposed NeRF (RAD-NeRF) framework as illustrated in Figure~\ref{fig:network}.
We first review the preliminaries of NeRF and the problem setting for audio-driven portrait synthesis (Section~\ref{sec:prelim}).
Then we elaborate the Decomposed Audio-spatial Encoding Module (Section~\ref{sec:head}) and the Pseudo-3D Deformable Module (Section~\ref{sec:torso}) for head and torso modeling, respectively.
Finally, we describe the essential training details that are distinguished from previous methods (Section~\ref{sec:details}).

\subsection{Preliminaries}
\label{sec:prelim}

\noindent \textbf{Neural Radiance Fields.}
NeRF~\cite{mildenhall2020nerf} is proposed to represent a static 3D volumetric scene with a 5D plenoptic function $\mathcal F: \mathbf{x}, \mathbf{d} \rightarrow \sigma, \mathbf{c}$, where $\mathbf{x} = (x,y,z)$ is the 3D coordinate, $\mathbf{d} = (\theta, \phi)$ is the viewing direction, $\sigma$ is the volume density, and $\mathbf{c} = (r, g, b)$ is the emitted color.
Given a ray $\mathbf{r}$ originating from $\mathbf{o}$ with direction $\mathbf{d}$, we query $\mathcal F$ at points $\mathbf{x}_i = \mathbf{o} + t_i \mathbf{d}$ sequentially sampled along the ray for computing densities $\{\sigma_i\}$ and colors $\{\mathbf{c}_i\}$.
The color of the pixel corresponding to the ray is obtained by numerical quadrature:
\begin{equation}
\label{eq:volume_rendering}
    \mathbf{\hat C}(\mathbf{r}) = \sum_i T_i \alpha_i \mathbf{c}_i;
    ~~T_i = \prod_{j < i} (1 - \alpha_j), 
\end{equation}
where $T_i$ is the transmittance, $\alpha_i = 1 - \exp(-\sigma_i \delta_i)$ is the opacity, and $\delta_i = t_{i+1} - t_i$ is the step size.
Based on this fully differentiable volume rendering procedure, NeRF can learn 3D scenes with supervision from only 2D images.

\noindent \textbf{Dynamic NeRF.}
In terms of dynamic scenes novel view synthesis, an additional condition (\textit{i.e.}, the current time $t$) is required. Previous methods usually perform dynamic scene modeling via two approaches:
1) \textit{Deformation-based} methods~\cite{pumarola2021d,park2021nerfies} learn a deformation $\Delta \mathbf{x}$ at each position and time step: $\mathcal G: \mathbf{x}, t \rightarrow \Delta \mathbf{x}$, which is subsequently added to the original position $\mathbf{x}$.
2) \textit{Modulation-based} methods~\cite{guo2021ad,gafni2021dynamic} directly condition the plenoptic function on time: $\mathcal F: \mathbf{x}, \mathbf{d}, t \rightarrow \sigma, \mathbf{c}$.

Since the deformation-based methods are not good at modeling topological changes (\textit{e.g.}, mouth openings and closings) due to the intrinsic continuity of the deformation field~\cite{park2021hypernerf}, we choose the modulation-based strategy to model the head part, and the deformation-based strategy to model the torso part with simpler motion patterns.

\noindent \textbf{Audio-driven Neural Radiance Talking Portrait.}
We briefly describe the common pipeline of audio-driven neural talking portrait synthesis methods~\cite{guo2021ad,shen2022learning,liu2022semantic}.
The training data is usually a 3-5 minute scene-specific video with a synchronized audio track, recorded by a static camera.
There are three main preprocessing steps for each image frame:
(1) Semantic parsing~\cite{lee2020maskgan} of the head, neck, torso, and background part;
(2) Extracting 2D facial landmarks~\cite{bulat2017far}, including eyes and lips;
(3) Face-tracking~\cite{thies2016face2face} to estimate the head pose parameters.
Note that these steps are only required for the training procedure.
For audio processing, an Automatic Speech Recognition (ASR) model~\cite{amodei2016deep,baevski2020wav2vec} is applied to extract audio features from the audio track.
Based on the head poses and audio conditions, a NeRF can be used to learn to synthesize the head part.
Since the torso part is not in the same coordinate system as the head part, it requires separate modeling, \textit{e.g.}, by another full NeRF~\cite{guo2021ad}.

\noindent \textbf{Grid-based NeRF.}
Recent grid-based NeRF~\cite{mueller2022instant,TensoRF,sun2021direct} encode 3D spatial information of static scenes with a 3D feature grid encoder $E^3_\text{spatial}$:
$\mathbf{f} = E^3_\text{spatial}(\mathbf{x})$, 
where $\mathbf{x} \in \mathbb{R}^3$ is a spatial coordinate, and $\mathbf{f}$ is the encoded spatial features.
Such feature grid encoders replace the MLP forwarding with cheaper linear interpolation to query spatial features, significantly enhancing the efficiency of both training and inference.
This makes it possible to achieve real-time rendering speed for static 3D scenes~\cite{mueller2022instant}.
We take this inspiration and extend it to encode high-dimensional audio-spatial information required by dynamic talking portrait synthesis.

\begin{figure}[t!]
    \centering
    \includegraphics[width=\linewidth]{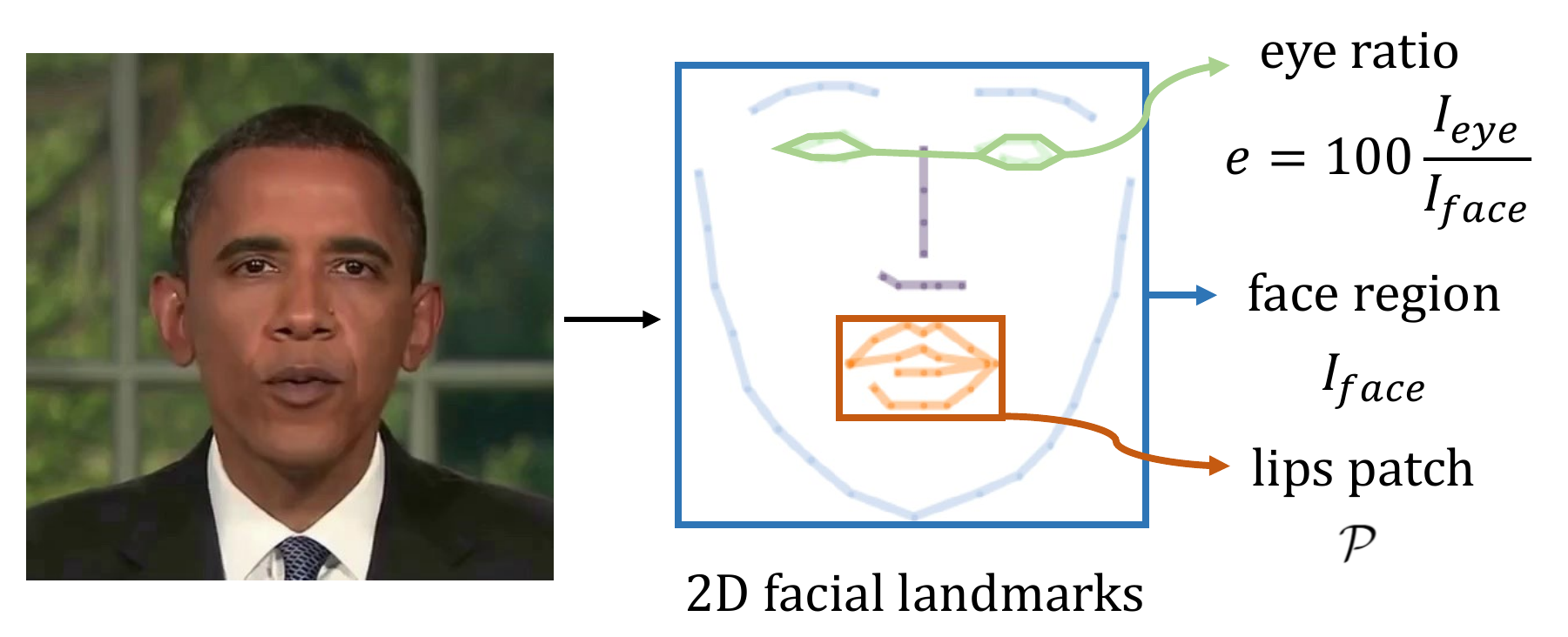}
    \caption{\textbf{An example of the landmark information}.
    Based on the predicted 2D facial landmarks, we extract three features to assist training: 
    the face region $\mathcal I_\text{face}$ for dynamic regularization, 
    the eye ratio $e$ for eye control, 
    and the lips patch $\mathcal P$ for lips fine-tuning.
    }
    \label{fig:landmark}
\end{figure}

\subsection{Decomposed Audio-spatial Encoding Module}
\label{sec:head}

\noindent \textbf{Audio-spatial Decomposition.}
Previous implicit NeRF methods~\cite{guo2021ad,liu2022semantic} usually encode audio signals into high-dimensional audio features and concatenate them with spatial features.
However, it's not trivial to integrate high-dimensional features with grid-based NeRF, since the complexity of linear interpolation grows exponentially as the input dimension increases.
It soon becomes computationally unaffordable if we directly use the high-dimensional concatenated audio-spatial features in a grid encoder.
Therefore, we propose two designs to \emph{alleviate the curse of dimensionality} for modeling audio-spatial information.

Firstly, we compress the high-dimensional audio features $\mathbf{a}$ into a low-dimensional audio coordinate $\mathbf{x}_a \in \mathbb R^D$, where the dimension $D \in [1, 2, 3]$ is small.
This is achieved through an MLP in a spatial-dependent manner: $\mathbf{x}_a = \text{MLP}(\mathbf{a}, \mathbf{f})$. 
We concatenate the spatial features $\mathbf{f}$ here so that the audio coordinate is explicitly dependent on the spatial position.
This operation relieves the audio features from implicitly learning the spatial information, which makes a more compact audio coordinate possible.
The audio coordinate is inspired by the deformable slicing surface type of ambient coordinate in HyperNeRF~\cite{park2021hypernerf}, but integrated with the feature grid encoder to achieve high efficiency.

Secondly, instead of using a composed audio-spatial grid encoder with higher dimension $\mathbf{g} = E^{3+D} (\mathbf{x}, \mathbf{x}_a)$, 
we decompose it into two grid encoders with lower dimensions to encode audio and spatial coordinates separately:
$\mathbf{f} = E^3_\text{spatial} (\mathbf{x}), \mathbf{g} = E^D_\text{audio} (\mathbf{x}_a)$.
This further reduces the interpolation cost from $2^{3 + D}$ to $2^3 + 2^D$ ($D \ge 1$).
The spatial features $\mathbf{f}$ and audio features $\mathbf{g}$ can be concatenated after performing interpolation.

\noindent \textbf{Explicit Eye Control.}
The eye movement is also a key factor for natural talking portrait synthesis.
However, since there is no strong correlation between eye blinks and audio signals, previous methods~\cite{guo2021ad, shen2022learning} often ignore the control of eyes, which leads to artifacts like too fast or half blink.
We provide a method to explicitly control the eye blinks.
As illustrated in Figure~\ref{fig:landmark}, we calculate the percentage of the eye area in the whole image based on the 2D facial landmarks, and use this ratio, which usually ranges from 0\% to 0.5\%,  as a one-dimensional eye feature $e$.
We condition the NeRF network on this eye feature, and show that this simple modification is enough for the model to learn the eye dynamics through the plain RGB loss.
At test time, we can easily adjust the eye percentage to control the eye blinks.

\noindent \textbf{Overall Head Representation.}
Concatenating the spatial features $\mathbf{f}$, audio features $\mathbf{g}$, eye feature $e$, along with a latent appearance embedding $\mathbf{i}$~\cite{martin2021nerf,gafni2021dynamic}, 
a small MLP is used to produce the density and color:
\begin{equation}
\mathbf{c}, \sigma = \text{MLP}(\mathbf{f}, \mathbf{g}, e, \mathbf{i})
\end{equation}
Volume rendering (Equation~\ref{eq:volume_rendering}) can then be applied to evaluate the pixel color and synthesize each head image frame.

\subsection{Pseudo-3D Deformable Module}
\label{sec:torso}

In contrast to the head part, the torso part is nearly static, only containing slight motions without topological change.
Previous methods either use another full dynamic NeRF to model the torso~\cite{guo2021ad}, or learn an entangled deformation field together with the head~\cite{liu2022semantic}. 
We consider these methods excessive, and propose a more efficient Pseudo-3D Deformable Module, as shown in the lower part of Figure~\ref{fig:network}.

Our method can be viewed as a 2D version of the deformation-based dynamic NeRF. 
Instead of sampling a series of points along each camera ray, we only need to sample one pixel coordinate $\mathbf{x}_t \in \mathbb{R}^2$ from the image space.
The deformation is conditioned on the head pose $\mathbf{p}$ such that the torso motion is synchronized with the head motion.
We adopt an MLP to predict the deformation: $\Delta \mathbf{x} = \text{MLP}(\mathbf{x}_t, \mathbf{p})$. 
The deformed coordinate is fed to a 2D feature grid encoder to get the torso feature: 
$\mathbf{f}_t = E^2_\text{torso} (\mathbf{x}_t + \Delta \mathbf{x})$.
Another MLP is used to produce the torso RGB color and alpha values: 
\begin{equation}
\mathbf{c}_t, \alpha_t = \text{MLP}(\mathbf{f}_t, \mathbf{i}_t)
\end{equation}
where $\mathbf{i}_t$ is a latent appearance embedding to introduce more model capacity. 

We show that this deformation-based module can successfully model the torso dynamics and synthesize natural torso images matching the head.
More importantly, the pseudo-3D representation through 2D feature grid is very lightweight and efficient. 
The separately rendered head and torso images can be alpha-composited with any provided background image to get the final output portrait image.

\subsection{Training Details}
\label{sec:details}

\noindent \textbf{Maximum Occupancy Grid Pruning.}
A common technique to improve NeRF's efficiency is to maintain an occupancy grid to prune the ray sampling space~\cite{mueller2022instant,TensoRF}.
This is straightforward for static scenes.
Since the occupancy grid is also static, we can use a 3D grid to store it.
For dynamic scenes, the occupancy value is also dependent on the dynamic conditions, which require additional dimensions. 
However, a higher-dimensional occupancy grid is more difficult to store and maintain, leading to greatly increased model size and training time.

We observe that for talking heads, the occupancy variation caused by changing audio conditions is usually small and negligible.
Therefore, instead of maintaining an occupancy grid for each audio condition, we propose to maintain a maximum occupancy grid for all audio conditions. 
In training, we randomly sample audio conditions from the training dataset and keep the maximum occupancy values.
In this way, we only need a 3D grid to store the occupancy values, and can successfully prune the ray sampling space under different audio conditions.

\noindent \textbf{Loss function.}
We use the MSE loss on each pixel's color $\mathbf C$ to train our network as the vanilla NeRF~\cite{mildenhall2020nerf}:
\begin{equation}
\label{eq:mse}
    \mathcal{L}_\text{color} = \sum_{\mathbf C \in \mathcal I} || \mathbf C - \mathbf C_\text{gt}  ||^2_2
\end{equation}
Besides, an entropy regularization term is used to encourage the pixel transparency to be either 0 or 1:
\begin{equation}
    \mathcal{L}_\text{entropy} = - \sum_{\alpha \in \mathcal I} (\alpha \log \alpha + (1 - \alpha) \log (1 - \alpha) )
\end{equation}
where $\alpha$ is the transparency for each pixel in image $\mathcal I$.

Ideally, the audio condition should only affect the facial region. 
To stabilize the dynamic modeling, we also propose an L1 regularization term on the audio coordinate:
\begin{equation}
    \mathcal{L}_\text{dynamic} = \sum_{\mathbf{x}_a \in \bar{\mathcal{I}}_\text{face}} | \mathbf{x}_a |
\end{equation}
This term encourages the audio coordinate $\mathbf{x}_a$ to be close to 0 at the non-facial region $\bar{\mathcal{I}}_\text{face}$, which helps to avoid unexpected flickerings outside the facial region, such as the hair and ears.
The facial region can be directly located by the 2D landmarks, as illustrated in Figure~\ref{fig:landmark}.

\noindent \textbf{Fine-tuning of the Lips.}
High-quality lips are of vital importance to make the synthesized portrait natural.
We found that the complex structural information of lips is hard to learn only through the pixel-wise MSE loss in Equation~\ref{eq:mse}.
Therefore, we propose to fine-tune the lips regions with patch-wise structural losses, \textit{e.g.}, the LPIPS~\cite{zhang2018perceptual} loss.
Instead of randomly sampling pixels from the whole image as the common NeRF training pipeline, we sample an image patch $\mathcal P$ where the lips are located based on the facial landmarks.
Then we can apply a combination of the LPIPS loss with the MSE loss balanced by $\lambda$ to fine-tune the lips region:
\begin{equation}
    \mathcal{L}_\text{fine-tune} = \sum_{\mathbf C \in \mathcal P} || \mathbf C - \mathbf C_\text{gt}  ||^2_2 + \lambda ~\text{LPIPS}(\mathcal P, \mathcal P_\text{gt})
\end{equation}

%% file: experiments.tex
\begin{figure*}[t!]
    \centering
    \includegraphics[width=\textwidth]{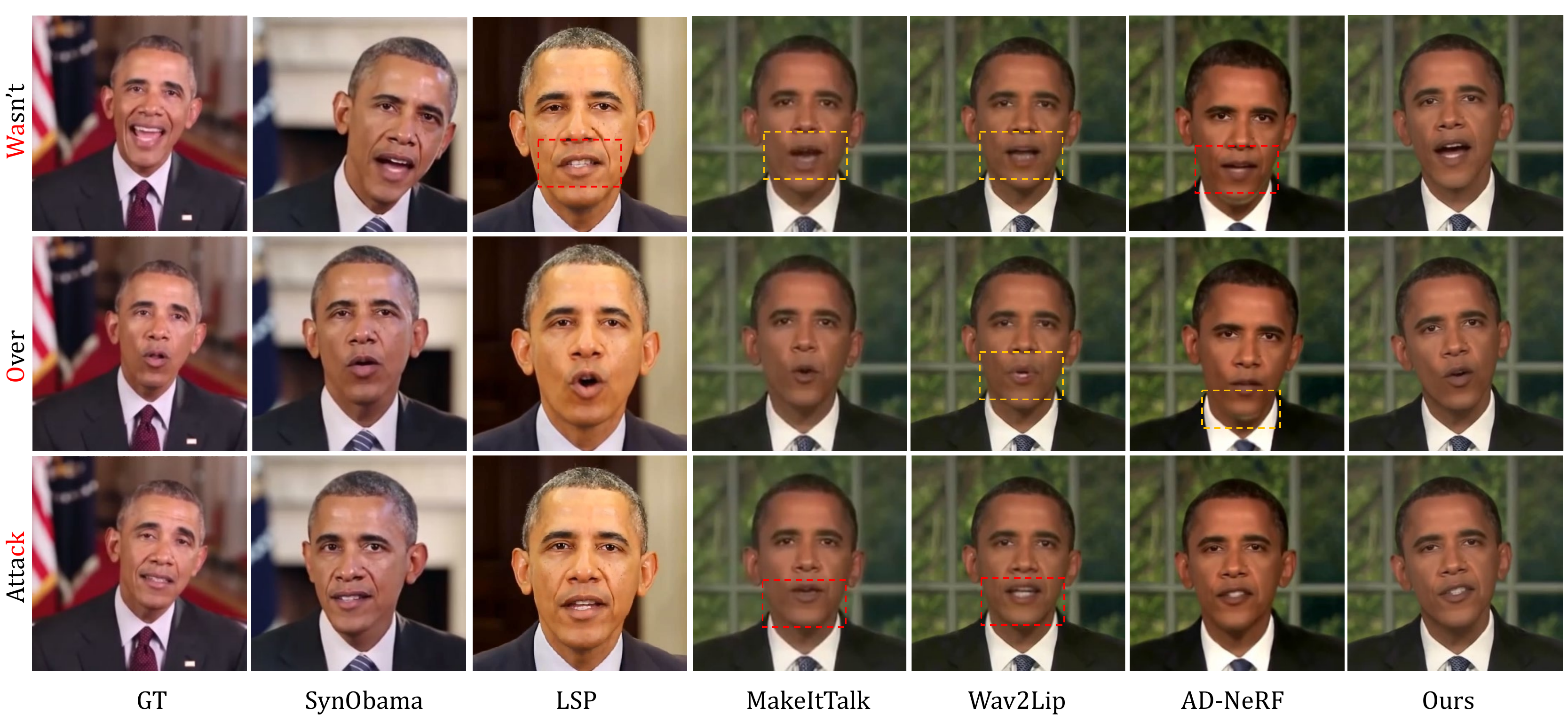}
    \caption{\textbf{Cross-driven quality comparison.}
    We show visualizations of representative methods under the cross-driven setting.
    Yellow boxes denote low image quality, and red boxes denote inaccurate lips.
    Our methods generate both good image quality and accurate lips movement.
    We recommend watching the \textcolor{violet}{supplementary video} for better details.
    }
    \label{fig:crossdriven}
\end{figure*}

\begin{figure}[t]
    \centering
    \includegraphics[width=\linewidth]{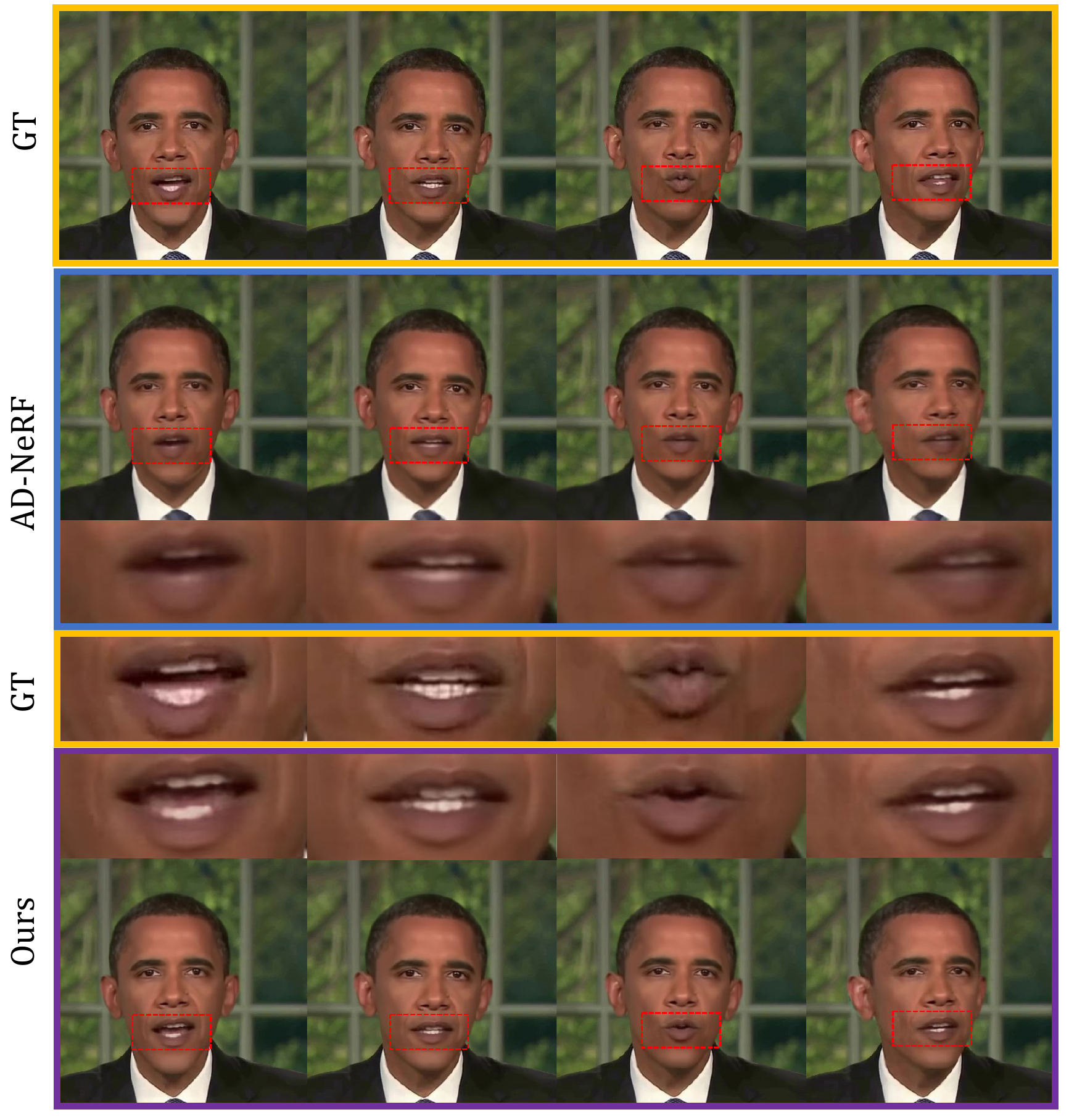}
    \caption{\textbf{Self-driven quality comparison.}
    We compare against the ground truth images under the self-driven setting. 
    Our method reconstructs sharper and more accurate lips compared to previous works.
    }
    \label{fig:selfdriven}
\end{figure}

\input{tabs/selfdriven_long.tex}

\subsection{Implementation Details}
We carry out experiments on the datasets collected by previous works~\cite{guo2021ad,shen2022learning}.
For each person-specific dataset, 
we train the head part for $20,000$ steps and fine-tune lips for $5,000$ steps, with each step containing $256^2$ rays. 
At most $16$ points are sampled per ray, pruned by the maximum occupancy grid.
The loss scale is set to $0.001$ for $\mathcal{L}_\text{entropy}$, $0.1$ for $\mathcal{L}_\text{dynamic}$, and $0.01$ for the LPIPS loss.
The Adam optimizer~\cite{kingma2014adam} with an initial learning rate of $0.0005$ for the network and $0.005$ for the feature grid is used.
The final learning rates are exponentially decayed to $0.1$ of the initial values.
We also use an Exponential Moving Average (EMA) of $0.95$ to make training more stable.
After the head part is converged, we train the torso part for $20,000$ steps with the same learning rate strategy.
The training for the head and torso takes around 5 and 2 hours respectively.
All the experiments are performed on one NVIDIA V100 GPU.
Please refer to the supplementary materials for more details.

\subsection{Quantitative Evaluation}

\noindent \textbf{Comparison Settings and Metrics.}
We evaluate our method under two settings, self-driven and cross-driven.
For the self-driven setting, we use the Obama dataset from AD-NeRF~\cite{guo2021ad} with the same data split for training and testing.
Since we have the ground truth of the same identity, we can use metrics like {PSNR}, {LPIPS}~\cite{zhang2018perceptual}, and {LMD (Landmark distance)}~\cite{chen2018lip} to evaluate the portrait reconstruction quality.
We recommend the {LMD} metric which directly measures the distance between lips landmarks.
For the cross-driven setting, we choose two audio clips from NVP~\cite{thies2020neural} and SynObama~\cite{suwajanakorn2017synthesizing} to drive the other methods. 
In this case, we do not have the ground truth images of the same identity, so we follow previous works~\cite{guo2021ad,liu2022semantic} to adopt the identity agnostic {SyncNet confidence (Sync)}~\cite{chung2016out} and {Action Units Error (AUE)}~\cite{baltrusaitis2018openface} to measure the audio-lips synchronization and lips-related muscle action consistency.

\noindent \textbf{Evaluation Results.}
The self-driven evaluation results are shown in Table~\ref{tab:selfdriven}.
We compare against recent methods that can generate full-resolution video as the ground truth.
Our method achieves better quality in most metrics, with a real-time inference FPS.
In specific, our method infers about 500$\times$ faster compared to the baseline AD-NeRF, and also converges about 5$\times$ faster.
The cross-driven results are listed in Table~\ref{tab:crossdriven}.
We achieve comparable performance compared to recent representative methods, which demonstrate the accuracy of our synthesized lips.
Note that the SyncNet confidence of Wav2Lip~\cite{prajwal2020lip} can be better than the ground truth, since they directly use the pretrained SyncNet in a loss term.

\subsection{Qualitative Evaluation}

\begin{figure}[t]
    \centering
    \includegraphics[width=\linewidth]{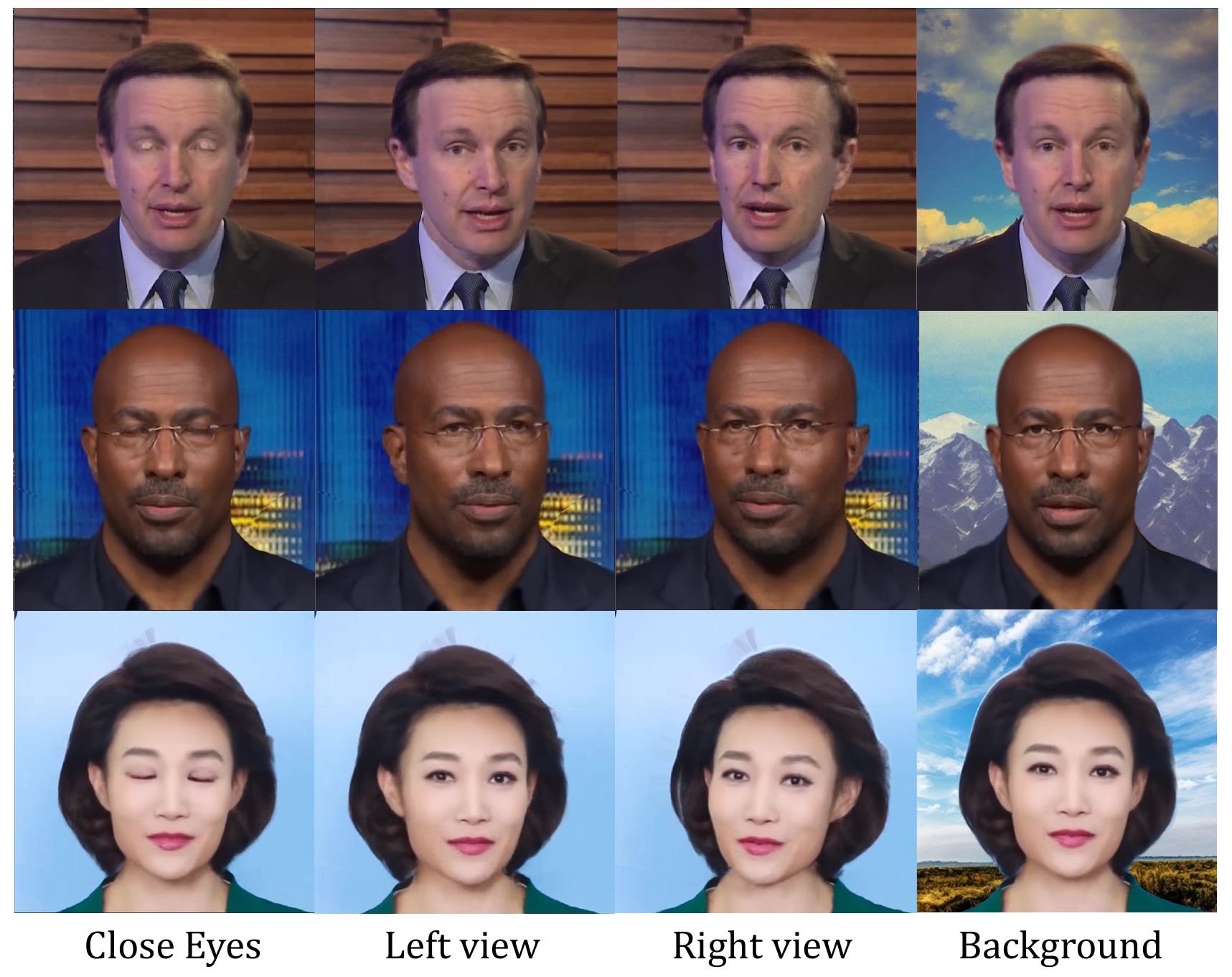}
    \caption{\textbf{Explicit control of talking portrait synthesis.}
    Apart from lips, our method also supports explicit control of eyes, head poses, and background images.
    }
    \label{fig:control}
\end{figure}

\noindent \textbf{Evaluation Results.}
In Figure~\ref{fig:selfdriven}, we show the visualization of our generated talking portraits under the self-driven setting.
The lips region is zoomed up for better details.
Our results are sharper and more accurate compared to AD-NeRF~\cite{guo2021ad}, while also being much faster in both training and inference.
In Figure~\ref{fig:crossdriven}, we compare the cross-driven setting video against representative methods.
We show that our method can robustly synthesize audio-synchronized lips, especially with large lips movement (\textit{e.g.}, the large-open mouth when pronouncing `wa').
Our results are comparable in quality to SynObama~\cite{suwajanakorn2017synthesizing}, which uses 14 hours of video to train, whereas we only train on a 5 minute video.
We also show the video manipulation ability of our method in Figure~\ref{fig:control}.

\input{tabs/crossdriven.tex}

\input{tabs/userstudy_long.tex}
\input{tabs/audio_dim.tex}
\input{tabs/backbone.tex}
\input{tabs/sampling_count.tex}

\noindent \textbf{User Study.}
To better evaluate the quality of generated talking faces, we conduct an extensive user study with the help of 20 attendees.
36 synthesized videos from different methods under the cross-driven setting are evaluated on three aspects:
audio-lips synchronization, image generation quality, and overall video realness.
Each aspect is rated between 1 and 5 (the larger the better), and the final results are summarized in Table~\ref{tab:userstudy}.
The user study further demonstrates the good quality of our generated talking portraits.

\subsection{Ablation Study}
We conduct an ablation study under the self-driven setting to verify the effect of each proposed module.

\noindent \textbf{Audio Coordinate Dimension.}
We first ablate the influence of audio coordinate dimension $D$ in Table~\ref{tab:audio_dim}.
Using a smaller dimension results in faster inference speed, but the limited audio encoding may harm the generation quality, while a larger dimension is slower and harder to converge.
We find a 2D audio coordinate best suits our setting.

\begin{figure}[t]
    \centering
    \includegraphics[width=\linewidth]{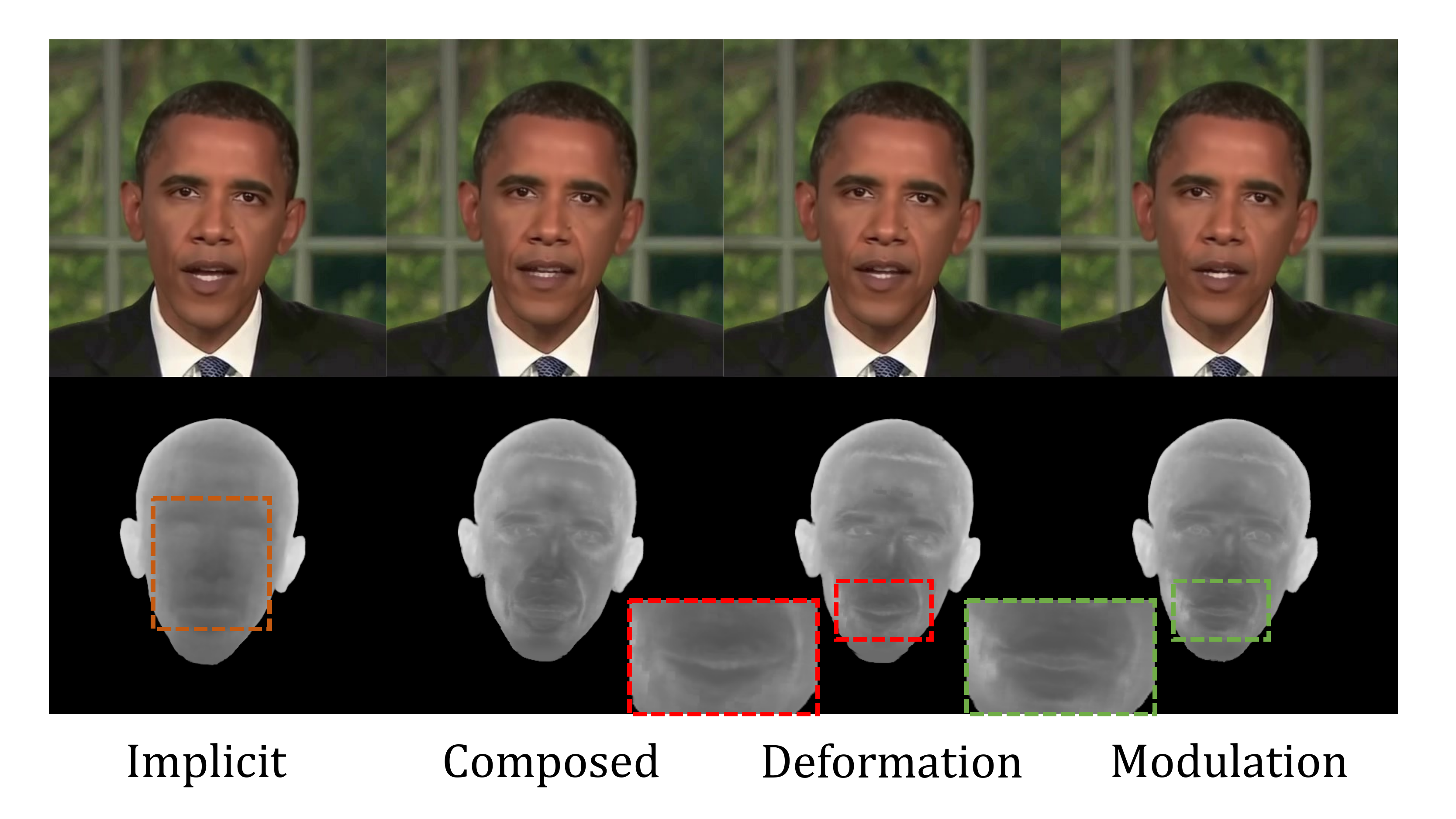}
    \caption{\textbf{Ablation on network backbone.}
    We show the visualization of RGB and Depth synthesized with different backbones.
    Implicit backbone tends to generate blurry depth, and deformation-based backbone fails to model topological changes in the lips region.
    }
    \label{fig:ablation_backbone}
    \vspace{-0.3cm}
\end{figure}

\noindent \textbf{Dynamic Modeling.}
In Table~\ref{tab:backbone} and Figure~\ref{fig:ablation_backbone}, we experiment with different settings of the backbone model for modeling head:
1) \textit{Implicit} backbone only using MLPs. We adopt a backbone network similar to AD-NeRF~\cite{guo2021ad}, but with our maximum occupancy grid pruning for acceleration. 
2) \textit{Composed} audio-spatial feature grid with smaller MLPs. A plain 5D grid encoder is used to encode the 5D audio-spatial coordinates.
3) Decomposed audio-spatial feature grid with smaller MLPs. We test both \textit{deformation}- and \textit{modulation}-based dynamic modeling strategies. 
We observe that the maximum occupancy grid based ray pruning already accelerates the implicit backbone to a significant level, but the synthesized depth is blurry.
With the proposed decomposed audio-spatial encoding, we achieve the best balance between performance and speed.

\noindent \textbf{Ray Sampling Count.}
We also conduct experiments on the max number of sampling points per ray in Table~\ref{tab:sample_cnt}.
We found that different from general 3D scenes~\cite{mildenhall2020nerf}, human heads are simpler and require fewer sampling points to reach good rendering quality. 
In our experiments, 16 sampling points per ray are enough to synthesize realistic images while keeping fast inference speed.

%% file: tabs/selfdriven_long.tex
\begin{table*}[t]
\begin{center}
\setlength{\tabcolsep}{10pt}
\renewcommand{\arraystretch}{1.1}
\begin{tabular}{l cc  ccc cc} 
\shline
Methods                              & PSNR$\uparrow$ & LPIPS$\downarrow$ & LMD$\downarrow$ & Sync$\uparrow$ & AUE$\downarrow$ & Training Time$\downarrow$ & Inference FPS$\uparrow$  \\   
Ground Truth                         & $\infty$   &  $0$  & $0$     & $8.448$          & $0$     & -             & -   \\
\hline
MakeItTalk~\cite{zhou2020makelttalk} & -       & -        & $4.484$ & $5.440$          & $1.327$ & -     & $12$            \\
Wav2Lip~\cite{prajwal2020lip}        & $30.94$ & $0.0639$ & $3.318$ & $\mathbf{7.756}$ & $1.120$ & -     & $15$            \\
AD-NeRF~\cite{guo2021ad}             & $25.53$ & $0.0901$ & $2.858$ & $5.428$          & $1.043$ & $36$h & $0.08$          \\
\hline
Ours                                 & $\mathbf{34.00}$ & $\mathbf{0.0387}$ & $\mathbf{2.696}$ & $6.664$ & $\mathbf{0.882}$ & $\mathbf{7}$h  & $\mathbf{40}$ \\
\shline 
\end{tabular}
\end{center}
\vspace{-0.3cm}
\caption{\textbf{Quantitative comparison under the self-driven setting.} 
We perform self-driven synthesis on the same identity's testset and compare the face reconstruction quality.
MakeItTalk~\cite{zhou2020makelttalk} cannot generate the same head poses as the ground truth video, so the PSNR and LPIPS are not reported.
The training time is only reported for person-specific methods.
}
\label{tab:selfdriven}
\end{table*}

%% file: tabs/crossdriven.tex
\begin{table}[t]
\begin{center}
\small
\renewcommand{\arraystretch}{1.1}
\begin{tabular}{l c c c c} 
\shline
        & \multicolumn{2}{c}{Testset A~\cite{suwajanakorn2017synthesizing}}  & \multicolumn{2}{c}{Testset B~\cite{thies2020neural}} \\ 
\hline
Methods & Sync$\uparrow$ & AUE$\downarrow$ & Sync$\uparrow$ & AUE$\downarrow$     \\   
Ground Truth                                 & $7.999$ & $0    $  & $6.913$ & $0    $   \\
\hline
MakeItTalk~\cite{zhou2020makelttalk}         & $5.287$ & $1.722$  & $5.144$ & $1.359$   \\
Wav2Lip~\cite{prajwal2020lip}                & $9.257$ & $1.904$  & $8.565$ & $1.493$   \\
SynObama~\cite{suwajanakorn2017synthesizing} & $7.297$ & $1.949$  & $-    $ & $-    $   \\
NVP~\cite{thies2020neural}                   & $-    $ & $-    $  & $3.476$ & $1.110$   \\
LSP~\cite{lu2021live}                        & $5.623$ & $1.895$  & $5.732$ & $1.363$   \\
AD-NeRF~\cite{guo2021ad}                     & $4.455$ & $2.133$  & $4.614$ & $1.801$   \\
\hline
Ours                                         & $6.218$ & $1.854$  & $5.500$ & $1.584$   \\
\shline 
\end{tabular}
\end{center}
\vspace{-0.3cm}
\caption{\textbf{Quantitative Comparison under the cross-driven setting.} 
We extract two audio clips from SynObama~\cite{suwajanakorn2017synthesizing} and NVP~\cite{thies2020neural} to drive the other methods, and compare the audio-lips synchronization and lips movement consistency.
}
\label{tab:crossdriven}
\end{table}

%% file: tabs/userstudy_long.tex
\begin{table*}[t]
\begin{center}
\small
\renewcommand{\arraystretch}{1.1}
\begin{tabular}{l c c c c c c c c} 
\shline
Methods & NVP~\cite{thies2020neural} & SynObama~\cite{suwajanakorn2017synthesizing} 
        & Wav2Lip~\cite{prajwal2020lip} & MakeItTalk~\cite{zhou2020makelttalk} 
        & LSP~\cite{lu2021live} & AD-NeRF~\cite{guo2021ad} & Ours \\   
\hline
Audio-Lips Sync & $3.29$ & $\mathbf{4.10}$ & $3.26$ & $2.33$ & $3.69$ & $2.62$ & $\underline{3.74}$ \\
Image Quality & $3.24$ & $\underline{3.86}$ & $1.62$ & $1.83$ & $3.74$ & $2.60$ & $\mathbf{4.00}$ \\
Video Realness & $3.43$ & $\mathbf{3.90}$ & $1.40$ & $1.48$ & $3.26$ & $1.93$ & $\underline{3.55}$ \\
\shline 
\end{tabular}
\end{center}
\vspace{-0.3cm}
\caption{\textbf{User study.} 
The rating is of scale 1-5, the higher the better.
We emphasize the \textbf{best} and \underline{second best} results.
}
\label{tab:userstudy}
\end{table*}

%% file: tabs/audio_dim.tex
\begin{table}[t]
\begin{center}
\small
\renewcommand{\arraystretch}{1.1}
\begin{tabular}{l c c c c} 
\shline
Audio Dimension & FPS$\uparrow$ & PSNR$\uparrow$ & LPIPS$\downarrow$ & LMD$\downarrow$ \\ 
\hline
$1$ & $40$ & $33.99$ & $0.0376$ & $2.736$ \\
$2$ & $40$ & $34.07$ & $0.0375$ & $2.696$ \\
$3$ & $36$ & $34.16$ & $0.0377$ & $2.749$ \\
\shline 
\end{tabular}
\end{center}
\vspace{-0.3cm}
\caption{\textbf{Audio Coordinate Dimension.} 
Comparison of the inference FPS and image quality for different audio coordinate dimensions.
}
\label{tab:audio_dim}
\end{table}

%% file: tabs/backbone.tex
\begin{table}[t]
\begin{center}
\small
\centering\setlength{\tabcolsep}{8pt}
\renewcommand{\arraystretch}{1.1}
\begin{tabular}{l c c c c } 
\shline
Backbone & FPS$\uparrow$ & PSNR$\uparrow$ & LPIPS$\downarrow$ & LMD$\downarrow$ \\ 
\hline
Implicit                 & $28$ & $34.01$ & $0.0324$ & $2.701$ \\
Composed                 & $35$ & $34.22$ & $0.0371$ & $2.707$ \\
Deformation              & $39$ & $33.43$ & $0.0382$ & $3.225$ \\
Modulation               & $40$ & $34.07$ & $0.0375$ & $2.696$ \\
\shline 
\end{tabular}
\end{center}
\vspace{-0.3cm}
\caption{\textbf{Dynamic Modeling Backbone.} 
Comparison of the inference FPS and image quality for different backbones to model head dynamics.
}
\label{tab:backbone}
\end{table}

%% file: tabs/sampling_count.tex
\begin{table}[t]
\begin{center}
\small
\renewcommand{\arraystretch}{1.1}
\begin{tabular}{l c c c c} 
\shline
Sampling Count & FPS$\uparrow$ & PSNR$\uparrow$ & LPIPS$\downarrow$ & LMD$\downarrow$ \\ 
\hline
$8 $  & $51$ & $33.97$ & $0.0363$ & $2.761$ \\
$16$  & $40$ & $34.07$ & $0.0375$ & $2.696$ \\
$32$  & $35$ & $34.13$ & $0.0375$ & $2.661$ \\
\shline 
\end{tabular}
\end{center}
\vspace{-0.3cm}
\caption{\textbf{Ray Sampling Count.} 
Comparison of the inference FPS and image quality for different audio coordinate dimensions.
}
\label{tab:sample_cnt}
\end{table}

%% file: impact.tex
\noindent \textbf{Ethical Considerations.}
Our method can synthesize realistic talking portrait videos, given a short video of a specific person for training.
However, this technique could bring potential misuse issues.
As part of our responsibility, we are committed to combating malicious uses and supporting the Deepfake detection community.
We will share our generated videos to improve the robustness of fake video detection methods.
We believe that the appropriate use of our method will be helpful to the healthy development of digital human technology.

\noindent \textbf{Limitations and Future Work.}
A common problem for NeRF-based talking portrait synthesis methods~\cite{guo2021ad,liu2022semantic} is that accurate face semantic parsing is needed to separate the head and torso parts. 
Current methods fail to synthesize natural videos for persons with shoulder-length hair, but this problem can be avoided in a static head pose setting.
Also, since we rely on English ASR models~\cite{amodei2016deep,baevski2020wav2vec} to extract audio features, the lips could be less accurate when guided with a different language.
Audio-to-phone models~\cite{li2020universal} could be used to alleviate the inconsistency between training and testing language.

%% file: conclusion.tex
In this paper, we present a novel approach for real-time audio-driven neural talking portrait synthesis.
We propose a decomposed audio-spatial encoding module for head modeling and a pseudo-3D deformable module for torso modeling. 
Our approach enables efficient inference which is 500 times faster than previous works with better rendering quality.
Extensive experiments and user studies are conducted to verify the enhanced talking portrait synthesis quality of our method.

%% file: appendix.tex
\begin{appendix}
\appendix

\section{Additional Training Details}

\subsection{Data Processing}

\begin{figure}[t]
    \centering
    \includegraphics[width=\linewidth]{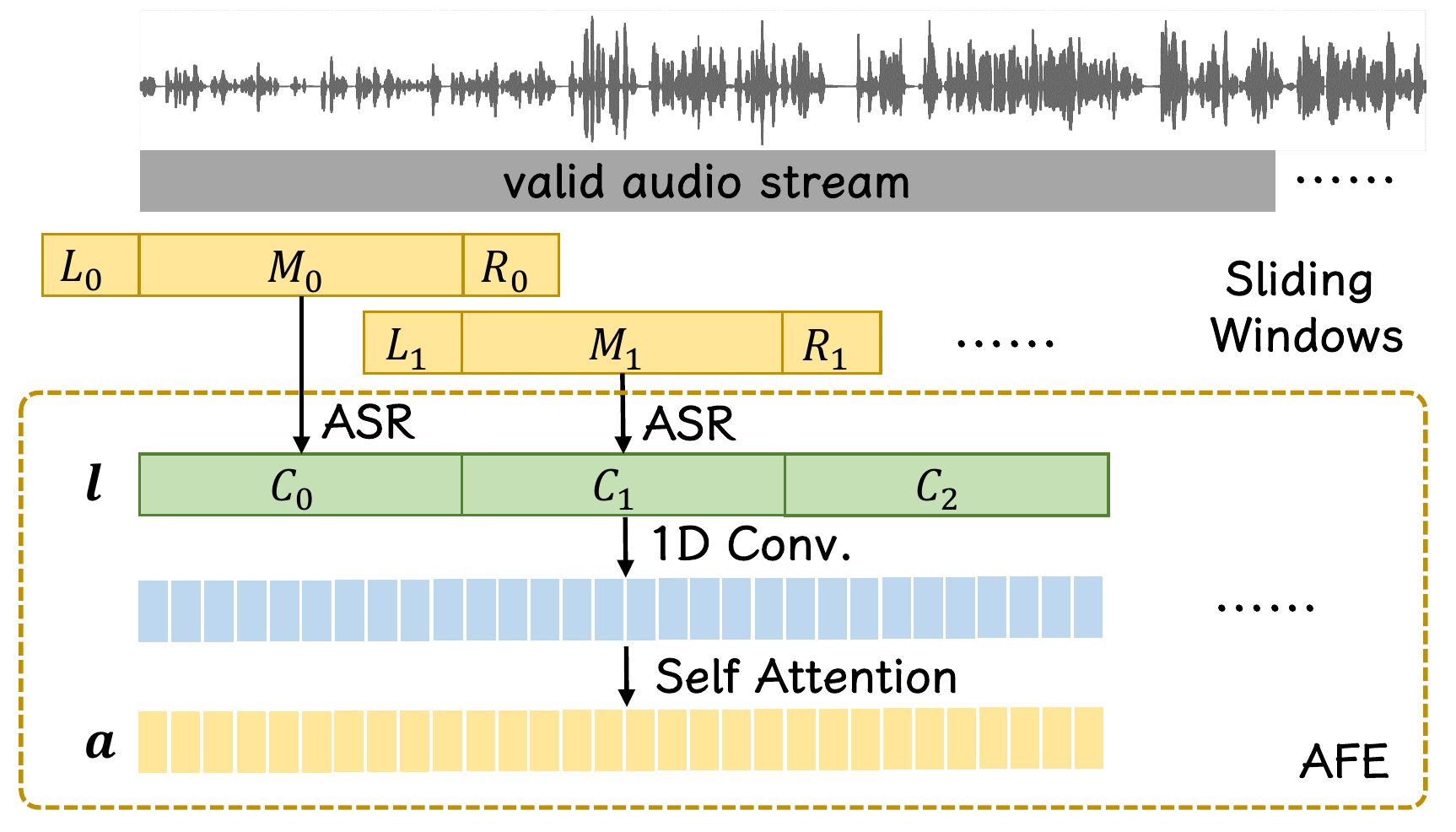}
    \caption{\textbf{Audio Feature Extractor}.
    We extract the audio features based on a sliding window strategy.
    An off-the-shelf ASR model is used to predict the classification logits $\mathbf{l}$ for each audio slice.
    The labels are temporally smoothed with 1D convolutions and a self-attention module to get the final audio features $\mathbf{a}$.
    }
    \label{fig:afe}
\end{figure}

\noindent \textbf{Audio Feature Extraction.}
Extracting information-rich audio features from input audio signals is a premise for accurate talking head synthesis.
Our Audio Feature Extractor (AFE) is shown in Figure~\ref{fig:afe}.
Following previous audio-driven methods~\cite{guo2021ad,thies2020neural}, we employ an off-the-shelf ASR model to predict the classification logits for each 20ms audio clip as the audio features.
Specifically, we adopt DeepSpeech~\cite{amodei2016deep} as the previous methods~\cite{guo2021ad,liu2022semantic} in experiments for fair comparisons.
We also found that other recent ASR models like Wav2Vec~\cite{baevski2020wav2vec} can lead to similar performance. 
This is as expected since the classification logits should be similar as long as the model can correctly recognize the audio signals.
These audio features are further filtered by an audio attention module following~\cite{guo2021ad}, which plays the role of temporal smoothing, to generate the final audio condition codes $\mathbf{a}$.

\noindent \textbf{Neck In-painting.}
To better learn the connections between the head and torso part, we use a neck in-painted torso image as the ground truth for torso training inspired by DFRF~\cite{shen2022learning}.
The in-painting algorithm empirically replicates the top pixels of the segmented neck part.
We further exponentially darken the pixel colors to simulate the shadow.
This helps to avoid head-torso disconnections, especially under cross-driven settings.

\subsection{Additional Implementation Details}

\noindent \textbf{Network Architecture.}
We report the detailed network architecture in this section.
The dimension for audio condition codes $\mathbf{a}$ is 64.
The audio feature extractor follows the implementation of AD-NeRF~\cite{guo2021ad}.
The window size for audio features is 16, and for the attention module is 8.
The spatial grid encoder $E_\text{spatial}$ uses 16 resolution levels starting from $16^3$ to $2048^3$, with each level containing 2 channels, and the hashmap size is limited to $2^{16}$.
The audio grid encoder $E_\text{audio}$ and torso grid encoder $E_\text{torso}$ share the same hyper-parameters as $E_\text{spatial}$ except the input dimension is set to 2.
The audio and density MLP both contain 3 layers with 64 hidden dimensions, and the color MLP contains 2 layers with 64 hidden dimensions.
Truncated exponential activation~\cite{mueller2022instant} is used for density, and sigmoid activation is used for color.

\noindent \textbf{Explicit Smoothing.}
A common problem in audio-driven talking head synthesis is the motion jitters in lips.
We propose a simple method to alleviate this artifact.
At test time, we apply momentum on the audio features: $\mathbf{\hat a}_{i} = \beta \mathbf{\hat a}_{i-1} + (1 - \beta) \mathbf{a}$ where $\beta$ is the smoothing hyper-parameter.
We found this effectively reduces the motion jitters and generates more natural videos.

\section{Additional Experimental Results}

\begin{figure}[t]
    \centering
    \includegraphics[width=\linewidth]{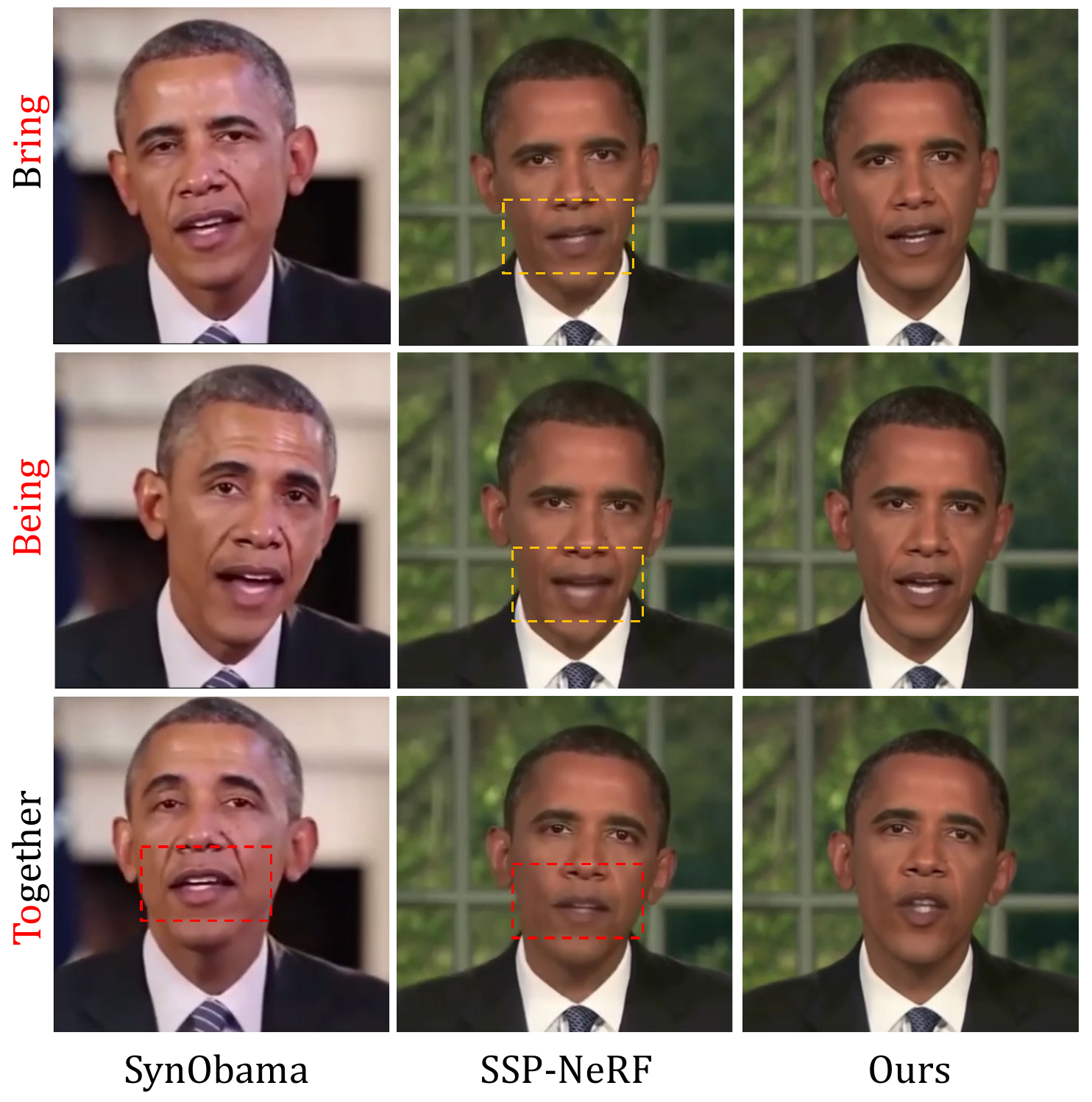}
    \caption{\textbf{More qualitative comparisons.}
    We compare our method with SynObama~\cite{suwajanakorn2017synthesizing} and SSP-NeRF~\cite{liu2022semantic}. Yellow boxes denote low image quality, and red boxes denote inaccurate lips.
    }
    \label{fig:quality_more}
\end{figure}

We show qualitative comparisons against SSP-NeRF~\cite{liu2020neural} in Figure~\ref{fig:quality_more}. 
Since there is no publicly available code for reproducing, we crop their best-quality demonstration video and use the corresponding audio to drive our method.
Our results show more sharp and accurate lips.
For more results, we recommend watching our \textcolor{violet}{supplementary video}.

\end{appendix}